\documentclass{article}

\usepackage{microtype}
\usepackage{graphicx}
\usepackage{subcaption}
\usepackage{booktabs} % for professional tables

\usepackage{hyperref}

\newcommand{\tb}[1]{\textbf{ #1 }}
\newcommand{\tg}[1]{\textcolor{lightgray}{ #1 }}

\usepackage[accepted]{icml_r2fm}

\usepackage{amsmath}
\usepackage{amssymb}
\usepackage{mathtools}
\usepackage{amsthm}

\usepackage[capitalize,noabbrev]{cleveref}

\theoremstyle{plain}

\theoremstyle{definition}

\theoremstyle{remark}

\usepackage[textsize=tiny]{todonotes}

\icmltitlerunning{What do Geometric Hallucination Detection Metrics Actually Measure?}

\begin{document}

\twocolumn[
\icmltitle{What do Geometric Hallucination Detection Metrics Actually Measure?}




\begin{icmlauthorlist}
\icmlauthor{Eric Yeats}{pnnl}
\icmlauthor{John Buckheit}{pnnl}
\icmlauthor{Sarah Scullen}{pnnl}
\icmlauthor{Brendan Kennedy}{pnnl}
\icmlauthor{Loc Truong}{pnnl}
\icmlauthor{Davis Brown}{pnnl,upenn}
\icmlauthor{Bill Kay}{pnnl}
\icmlauthor{Cliff Joslyn}{pnnl}
\icmlauthor{Tegan Emerson}{pnnl,csu,utep}
\icmlauthor{Michael J. Henry}{pnnl}
\icmlauthor{John Emanuello}{nsa}
\icmlauthor{Henry Kvinge}{pnnl,uw}
\end{icmlauthorlist}

\icmlaffiliation{pnnl}{Pacific Northwest National Laboratory}
\icmlaffiliation{uw}{University of Washington}
\icmlaffiliation{upenn}{University of Pennsylvania}
\icmlaffiliation{csu}{Colorado State University}
\icmlaffiliation{utep}{University of Texas, El Paso}
\icmlaffiliation{nsa}{Laboratory for Advanced Cybersecurity Research, National Security Agency}

\icmlcorrespondingauthor{Eric Yeats}{eric.yeats@pnnl.gov}

\icmlkeywords{Machine Learning, ICML}

\vskip 0.3in
]



\printAffiliationsAndNotice{}  % leave blank if no need to mention equal contribution

\begin{abstract}
Hallucination remains a barrier to deploying generative models in high-consequence applications. This is especially true in cases where external ground truth is not readily available to validate model outputs. This situation has motivated the study of geometric signals in the internal state of an LLM that are predictive of hallucination and require limited external knowledge. Given that there are a range of factors that can lead model output to be called a hallucination (e.g., irrelevance vs incoherence), in this paper we ask what specific properties of a hallucination these geometric statistics actually capture. To assess this, we generate a synthetic dataset which varies distinct properties of output associated with hallucination. This includes output \textit{correctness}, \textit{confidence}, \textit{relevance}, \textit{coherence}, and \textit{completeness}. We find that different geometric statistics capture different types of hallucinations. Along the way we show that many existing geometric detection methods have substantial sensitivity to shifts in task domain (e.g., math questions vs. history questions). Motivated by this, we introduce a simple normalization method to mitigate the effect of domain shift on geometric statistics, leading to AUROC gains of +34 points in multi-domain settings.
\end{abstract}

\section{Introduction}

Developing efficient and effective methods for detecting hallucinations is currently a major need for the larger goal of achieving reliable and responsible generative models \cite{huang2025survey}. Proposed approaches come in a range of flavors, from methods that validate large langauge model's (LLM's) output using external data sources \cite{lewis2020retrieval,asai2023self}, to methods that validate using a set of distinct judge LLMs \cite{jacovi2025facts,zheng2023judging}, to methods that use the consistency of a particular output as a signal for factual accuracy \cite{chen2024inside,manakul2023selfcheckgpt}. One family of methods that is particularly attractive is those that use signals extracted from model internals such as token representations in the residual stream or attention maps to detect when output is likely to be a hallucination. Though sometimes requiring labeled data upfront \cite{azaria2023internal,orgad2024llms}, these methods are often compute efficient to run and do not require external knowledge at inference time. Since model internals tend to be high-dimensional and not intrinsically interpretable, it has been common to leverage geometric or information-theoretic statistics in detection frameworks \cite{sriramanan2024llm,du2024haloscope,yin2024characterizing}.

However, hallucinations can come in many forms and be characterized in several ways \cite{huang2025survey}. What aspect of a hallucination a particular method is flagging remains mostly unexplored. 
In this paper we try to answer the question: \emph{what characteristics of a hallucination do popular geometric detection methods actually capture?} 
Specifically, we programmatically generate user prompts and model responses which exhibit different properties of a hallucination within different domains. We then run these prompts and responses through the model and extract hidden states within a teacher forcing framework \cite{sutton1988learning}. As the model has no `awareness' that it did not actually generate these responses, we can take these hidden states as a reasonable surrogate for natural model hallucinations.

Surprisingly, we find that different geometric properties of hidden states tend to correlate to varying degrees with different flavors of hallucinations. For instance, hidden score and attention score are sensitive to irrelevant responses while matrix entropy is sensitive to incoherent responses. This suggests that existing taxonomies of hallucination may be describable in the geometry of their representations. In the process of running these experiments, we note another important factor impacting detection effectiveness: the question domain (aligning with prior work on probes in \cite{liu2024universal}). For example, when detector evaluations involve multiple domains, the statistic variance across domains is significantly larger than the detection margin, harming performance. We introduce a novel normalization method to mitigate the impact of domain shift, leading to detector AUROC increases of $+34$ points in multi-domain settings. In summary, our work provides the following contributions: (1) We introduce a dataset to study the response of geometric statistics to hallucinated responses of various types: \textit{(factual) incorrectness}, \textit{(verbal) confidence}, \textit{irrelevance}, \textit{incoherence}, and \textit{incompleteness}. (2) We analyze the responses of geometric statistics to hallucinations of various types and severities across three domains. In the process of doing this we identify domain shift as a key challenge for these metrics. (3) To address this we propose a simple normalization method to mitigate the effect of domain shift on geometric statistics, leading to a $+34$ point increases in AUROC when detecting hallucinations across domains.

\section{Geometric Hallucination Detection Metrics}

Hallucination detectors that utilize the internals of an LLM tend to either operate on the token representation sequence in the residual stream or on the attention maps extracted from multihead attention. Here we review the three such methods that we use in the experiments in \cref{sec:initial_experiment}.

\textbf{Hidden Score:} Let $\mathbf{H}_l$ be the $m \times d$ matrix formed by the sequence of $d$-dimensional hidden representations of $m$ tokens following layer $l$ in the LLM residual stream. Let $\mathbf{G}_l:=\mathbf{H}_l\mathbf{H}_l^{T}$ be the corresponding Gram matrix for $\mathbf{H}_l$. \cite{sriramanan2024llm} defined the hidden score at layer $l$ to be the sequence-normalized log determinant of $\mathbf{G}_l$:
\begin{equation}
    \text{HS}(\mathbf{H}_l)=\frac{1}{m}\log\det\left(\mathbf{G}_l\right) = \frac{1}{m} \sum^m_{i=1}\log \lambda_i,
\end{equation}\label{eqn:hidden_score}
where $\lambda_i$ is the $i$-th eigenvalue of $\mathbf{G}_l$. The hidden score can be interpreted as the log volume of the parallelpiped formed by columns of $\mathbf{H}_l$. Hallucinations are associated with larger volumes, indicating a more diffuse representation.

Though it has not been used specifically to detect hallucinations, the matrix entropy (ME) of $\mathbf{G}_l$ was used in \cite{skean2025layer} to study the information content of LLM internal representations and is a natural statistic to characterize a sequence of hidden activation vectors. Matrix entropy is parameterized by a non-negative real-number $\alpha$. We use the version obtained by letting $\alpha \rightarrow 1$ which is equivalent to Shannon entropy\footnote{Matrix entropy uses the $\alpha$-Renyi entropy which generalizes several information-theoretic statistics including Shannon entropy. For instance, when $\alpha = 2$, $\alpha$-Renyi entropy corresponds to von Neumann entropy.}:

\begin{equation}
    \text{ME}(\mathbf{H}_l)=-\sum_{i=1}^m q_i\log q_i, \quad \quad q_i = \frac{\lambda_i}{\text{trace}(\mathbf{G}_l)}.
\end{equation}\label{eqn:matrix_entropy}
The matrix entropy provides another way of measuring how diffuse a sequence of token representations is.

Auto-regressive LLMs often employ causal self-attention. 
The corresponding attention maps are lower triangular and non-negative. Hence, the log determinant of the $i$-th attention map at layer $l$, $A^i_l$, is simply the sum of log entries along the diagonal. Introduced by \cite{sriramanan2024llm}, the attention score is the sequence-normalized log determinant for each of the $n$ attention heads:
\begin{equation}
    \text{AS}(A_l) = \frac{1}{m n} \sum_{i=1}^n \sum_{j=1}^m \log (A^i_l)_{j,j}.
\end{equation}\label{eqn:attention_score}
Informally, the attention score measures how much each token attends to itself. Experimentally, hallucinations are associated with higher attention scores.

\section{What do Geometric Statistics Measure?}\label{sec:initial_experiment}

We design an experiment to study the responses of geometric statistics to various hallucination types and levels of severity.

\paragraph{Dataset Design} Our dataset contains correct question and (numeric) answer (QA) pairs belonging to three domains: \texttt{math} (multiplication of positive integers), \texttt{history} (year of event, CE), and \texttt{counting} (number of times a word appears in a sequence). We use the QA pairs to generate prompt and response (PR) pairs. The PR pairs follow a template consisting of a \texttt{prompt\_question}, \texttt{response\_question}, \texttt{conf\_mod} (confidence modifier), \texttt{answer}, and \texttt{answer\_offset}. The template is:

\begin{center}
    \textbf{P:} ``\{\texttt{prompt\_question}\}'' \ \textcolor{lightgray}{``What is $46\times53$?''}

    \textbf{R:} ``The answer to `\{\texttt{response\_question}\}' is \{\texttt{conf\_mod}\} \{\texttt{answer} + \texttt{answer\_offset}\}.''

    \textcolor{lightgray}{``The answer to `What is $46 \times 53$?' is $2438$.''}
    
\end{center}

Given a QA pair (a \texttt{response\_question} and an \texttt{answer}), the baseline, correct PR sequence consists of a \texttt{prompt\_question} which is the \texttt{response\_question}, a \texttt{conf\_mod} set to ``'', and an \texttt{answer\_offset} set to $0$. We craft hallucinations by modifying the template in the following ways:

\textit{(Factual)} \textbf{\textit{Incorrectness}}: Assign a non-zero integer to \texttt{answer\_offset}, causing the response to be incorrect. Larger magnitude integers are used to simulate more severe hallucination.

\textit{(Under-)} \textbf{\textit{Confidence}:} Insert a verbal confidence modifier in place of \texttt{conf\_mod} e.g., ``\textit{probably}'' or ``\textit{maybe}''. 

\textbf{\textit{Irrelevance}:} Sample a \texttt{prompt\_question} which is different from \texttt{response\_question}. A cross-domain \texttt{prompt\_question} (as opposed to intra-domain) simulates a more severe hallucination.

\textbf{\textit{Incoherence}:} Repeat the response (\textbf{R}) multiple times while changing \texttt{answer\_offset}. More repetitions with inconsistent responses simulate stronger hallucinations.

\textbf{\textit{Incompleteness}:} Terminate the response (\textbf{R}) early with an end-of-text token. Earlier termination simulates more severe hallucination.

\paragraph{Data Recording} We collect the LLM's intermediate representations from each of the question and answer pairs using ``teacher forcing'' \cite{sutton1988learning}. The LLM ingests the PR pair, and the hidden states and attention maps are recorded for each token and LLM layer. We calculate and analyze HS, ME, and AS for each of these. We use Llama 3.1-8B-Instruct in our experiments \cite{grattafiori2024llama} which were run on a single Nvidia H100 GPU. In all cases, the task is detection of hallucinations (of any type and severity) among all responses, including correct ones.

\begin{table*}[ht]
\caption{Hallucination detection AUROC. Each reported AUROC is the highest score computed across all layers. The layer index $[0, 32)$ corresponding to the maximum AUROC is in parentheses ($\cdot$). We report (--) if multiple layers have the same AUROC.}
\label{tab:main_table}
\begin{center}
\begin{small}
\begin{sc}
\begin{tabular}{lccccccc}
\toprule
  & \multicolumn{3}{c}{\textit{incorrectness}} & \multicolumn{1}{c}{\textit{confidence}} & \multicolumn{1}{c}{\textit{irrelevance}} & \multicolumn{1}{c}{\textit{incoherence}} & \multicolumn{1}{c}{\textit{incompleteness}} \\
 & \small{Level 1} & \small{Level 2} & \small{Level 3} & \small{Level 3} & \small{Level 3} & \small{Level 3} & \small{Level 3} \\
\midrule
\textbf{\texttt{math}} & & & & & \\
HS & \tb{0.88} (30) & \tb{0.91} (30) & \tb{0.92} (30) & 0.99 (14) & 0.89 (00) & \tg{0.00 (--)} & 0.99 (--) \\
ME & 0.82 (30) & 0.90 (30) & \tb{0.92} (30) & 0.99 (--) & 0.99 (--) & 0.99 (--) & \tg{0.00 (--)} \\
AS & 0.71 (30) & 0.80 (31) & 0.86 (31) & 0.99 (21) & 0.98 (16) & \tg{0.00 (--)} & 0.99 (--) \\
\midrule
 \textbf{\texttt{history}} & & & & &\\
HS & \tb{0.66} (29) & 0.66 (16) & 0.69 (16) & 0.80 (14) & 0.98 (--) & \tg{0.00 (--)} & 0.97 (05) \\
ME & 0.54 (12) & 0.54 (16) & 0.56 (16) & 0.60 (14) & 0.76 (31) & 0.99 (--) & \tg{0.33 (31)} \\
AS & 0.61 (16) & \tb{0.70} (16) & \tb{0.75} (16) & 0.74 (16) & 0.99 (04) & \tg{0.00 (--)} & 0.92 (13) \\
\midrule
 \textbf{\texttt{counting}} & & & & &\\
HS & 0.51 (30) & 0.52 (30) & 0.53 (14) & 0.56 (15) & 0.99 (--) & \tg{0.00 (--)} & 0.86 (07) \\
ME & 0.52 (30) & 0.53 (30) & 0.53 (30) & 0.63 (00) & 0.95 (31) & 0.99 (--) & \tg{0.50 (31)} \\
AS & \tb{0.58} (16) & \tb{0.60} (16) & \tb{0.65} (16) & 0.79 (16) & 0.99 (05) & \tg{0.13 (01)} & 0.93 (17)\\
\midrule
 \textbf{\texttt{all}} & & & &\\
HS & 0.56 (30) & 0.56 (30) & 0.57 (30) & 0.69 (14) & 0.94 (00) & \tg{0.03 (30)} & 0.83 (12) \\
ME & 0.55 (30) & 0.56 (30) & 0.56 (30) & 0.59 (26) & 0.81 (31) & 1.00 (01) & \tg{0.39 (31)} \\
AS & 0.55 (31) & \tb{0.58} (30) & \tb{0.60} (30) & 0.66 (20) & 0.96 (05) & \tg{0.08 (00)} & 0.83 (17) \\
\bottomrule
\end{tabular}
\end{sc}
\end{small}
\end{center}
\end{table*}

\subsection{Experimental Results and Analysis}

\paragraph{Single-Domain Performance on \textit{Factual Incorrectness}} \cref{tab:main_table} depicts the area under the receiver operating characteristic (AUROC) hallucination detector scores for all statistics, domains, and hallucination types. We observe three primary trends regarding statistic responses to \textit{factual incorrectness}. \textbf{First}, more severe hallucinations (levels 1 to 3) are associated with better detection scores for all statistics. \textbf{Second}, different statistics are better detectors for different domains. Hidden Score (HS) and Matrix Entropy (ME) (those which process the Gram matrix of a hidden states sequence) are better detector scores for \texttt{math}, with AUROCS of 0.92 on level 3 hallucinations. However, Attention Score (AS) is the best detector score for \texttt{history} and \texttt{counting}, with AUROCs of 0.75 and 0.65 on level 3 hallucinations, respectively. \textbf{Third}, we observe that the domain impacts optimal recording layer more than the choice of metric. Layer indices 30-31 are the best for detecting hallucinations on the \texttt{math} dataset. However, layer indices 14-16 yield the highest AUROC for the three statistics on \texttt{history} and \texttt{counting}.

\paragraph{Sensitivity to Alternative Hallucination Types} The right-hand side of \cref{tab:main_table} reports AUROCs for the statistics as they respond to \textit{(under-)confidence}, \textit{irrelevance}, \textit{incoherence}, and \textit{incompleteness}. We observe that in general, the statistics are highly responsive to these types of hallucinations. For example, HS and AS achieve AUROCs of 0.94 and 0.96 for detecting \textit{irrelevant} hallucinations on \texttt{all}. Moreover, they yield AUROCs of 0.83 for \textit{incomplete} hallucinations on \texttt{all}. However, HS and AS do not respond to \textit{incoherent} responses as hallucinations (they have \textit{lower} scores than the baseline), leading to AUROCs far below random guessing for this type of hallucination. On the other hand, ME is highly receptive to \textit{incoherent} hallucinations, achieving near perfect AUROC for this category. Each of the statistics are somewhat responsive to \textit{under-confidence}, yielding AUROCs of 0.59-0.69 on \texttt{all}.

\begin{figure}[h]
    \begin{subfigure}{0.48\linewidth}
    \centering
        \includegraphics[width=\linewidth,trim=25pt 5pt 5pt 5pt, clip]{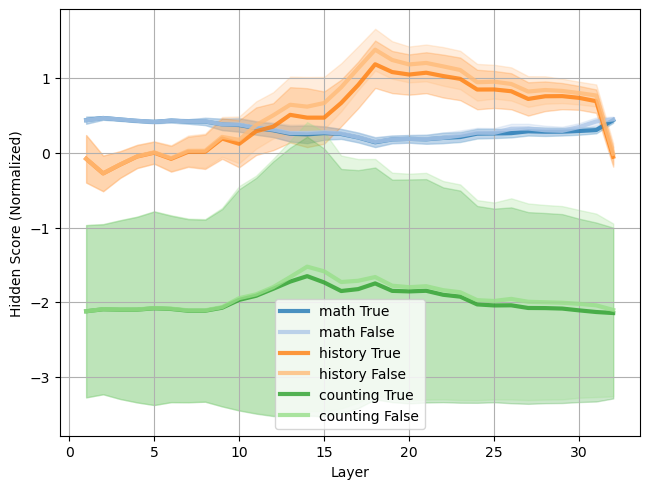}
    \end{subfigure}
    \hfill
    \begin{subfigure}{0.48\linewidth}
    \centering
        \includegraphics[width=\linewidth,trim=25pt 5pt 5pt 5pt, clip]{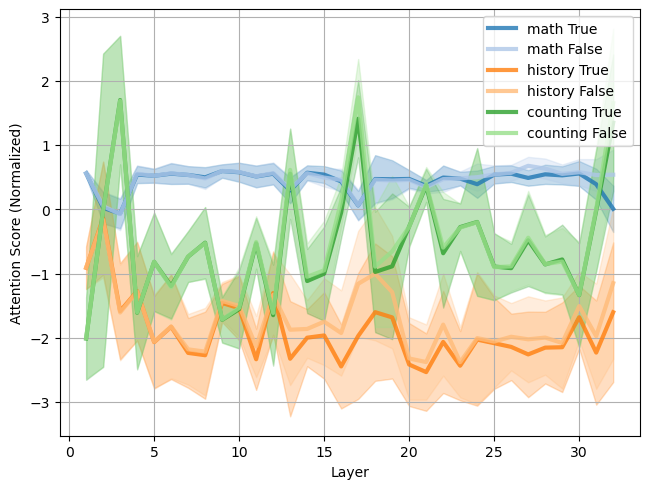}
    \end{subfigure}
    \caption{Distributions of HS (left) and AS (right) for \textit{correct} and \textit{incorrect} (level 3) responses for each domain. Solid lines: distribution means. Shaded areas: one standard deviation.}\label{fig:domain_comp}
\end{figure}

\paragraph{Domain Shift Harms Detection of \textit{Factual Incorrectness}} For \textit{factual incorrectness} questions and responses both HS and AS achieve strong AUROC on the individual domains (with the exception of HS on \texttt{counting}). However, when the domains are combined in \texttt{all}, the performance of each statistic drops significantly, with HS and AS yielding AUROCs of 0.57 and 0.60 on level 3 hallucinations. \cref{fig:domain_comp} depicts the distributions of HS and AS across domains. This poses issues in cases where one might want a detection system capable of catching hallucinations across domains (e.g., a general purpose chatbot).

\begin{figure*}[h!]
    \begin{subfigure}{0.31\linewidth}
    \centering
        \includegraphics[width=\linewidth,trim=25pt 5pt 5pt 5pt, clip]{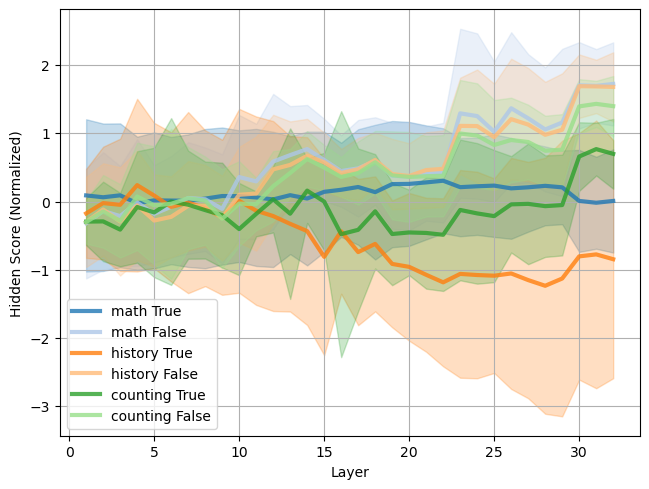}
        \caption{HS-Norm - Domain Alignment}\label{sfig:hs_diff_domain}
    \end{subfigure}
    \hfill
    \begin{subfigure}{0.31\linewidth}
    \centering
        \includegraphics[width=\linewidth,trim=25pt 5pt 5pt 5pt, clip]{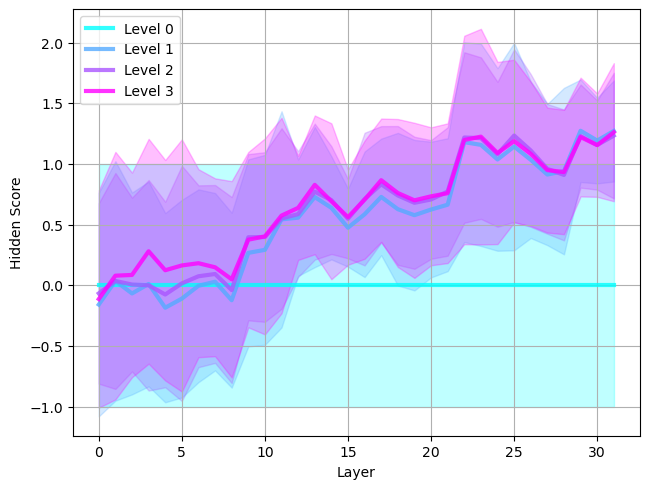}
        \caption{HS-Norm - score distribution on \texttt{all}}\label{sfig:hs_diff_all}
    \end{subfigure}
    \hfill
    \begin{subfigure}{0.31\linewidth}
    \centering
        \includegraphics[width=\linewidth,trim=5pt 5pt 5pt 5pt, clip]{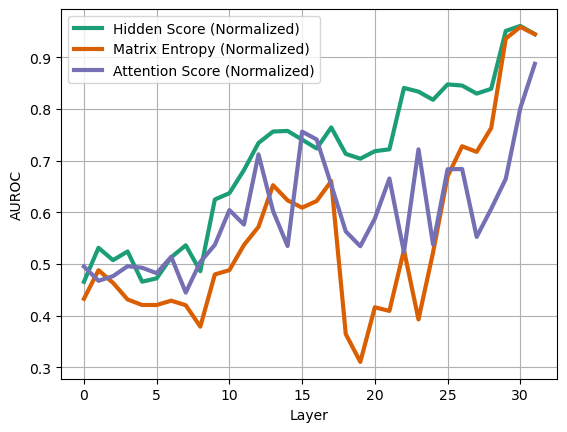}
        \caption{AUROC Comparison on \texttt{all} }\label{sfig:diff_auroc}
    \end{subfigure}
    \caption{Our \textit{perturbation normalization} significantly reduces domain shift, leading to boosted detection performance for \textit{incorrectness}.}\label{fig:domain_mitigation}
\end{figure*}

\section{Mitigating Domain Shift}

We propose a simple normalization method to mitigate domain shift, leading to improved detection performance. First, we locate the exact token(s) of the answer to be verified. This is simple in our experiments, but in real-world settings this task could be achieved by an auxiliary LLM. Second, we collect the representations of the original response $\mathbf{H}_l\in\mathcal{H}$ as well as $k$ versions of the same response $\mathbf{H}_l^1, ..., \mathbf{H}_l^k$ with \textit{perturbed} answers. This is done using a perturbation set suitable for the domain. In our setting we added $-5, -2, -1, 1, 2, 5$ to the responded answer. Third, for a given statistic function $f: \mathcal{H}_l \rightarrow \mathbb{R}$ and $\mu = \frac{1}{k}\sum^k_{i=1} f(\mathbf{H}_l^i)$, we compute the normalized score:

\begin{equation}
    f^*(\mathbf{H}_l)=\frac{ f(\mathbf{H}_l) - \mu}{\sqrt{\frac{1}{k}\sum^k_{i=1} 
    \left( f(\mathbf{H}_l^i) - \mu \right)^2}}.
\end{equation}

An equivalent definition holds when attention maps $A_l^1, \dots, A_l^n$ are used. Intuitively, $f^*(\mathbf{H}_l)$ measures how much of an outlier $f(\mathbf{H}_l)$ is compared with perturbed versions of itself. One should expect a \textit{correct} response to have lower statistics than its perturbed neighbors, while \textit{incorrect} responses should not. We call this \emph{perturbation normalization} of a geometric statistic.

\paragraph{Results}

We augment each of the statistics with $f^*(\cdot)$ and record their scores on \texttt{all}. \cref{sfig:hs_diff_domain} depicts the distributions of HS-Norm scores for baseline responses and for level-1 \textit{incorrectness} hallucinations (this should be compared with \cref{fig:domain_comp}). The \textit{correct} responses for each of the domains are aligned on the bottom half of the chart at layer 30, while the \textit{incorrect} response distributions are aligned on the top half of the chart at layer 30. \cref{sfig:hs_diff_all} depicts the normalized distributions of HS-Norm scores on \texttt{all} for \textit{correct} responses and \textit{incorrect} responses (levels 1-3). We observe that HS-Norm separates \textit{incorrect} from \textit{correct} well in the later stages of the network. \cref{sfig:diff_auroc} depicts the AUROC for each statistic on \texttt{all} (level 1 hallucinations). The aligned domains lead to significantly improved detector performance. HS-Norm and ME-Norm achieve AUROCs of 0.96 \textbf{(40 point gain)} at layer $30$, while AS-Norm achieves an AUROC of 0.89 \textbf{(34 point gain)} at layer $31$.

\section{Conclusion}

We design a multi-domain dataset of hallucinations with various properties and severities to try to answer the question ``What do geometric hallucination detection metrics actually measure?'' We find that all geometric statistics are correlated with \textit{incorrectness}, but different statistics respond to different hallucination characteristics. Additionally, we find that domain shift impairs the detection performance of the statistics on \textit{incorrectness}. We mitigate domain shift with a simple normalization technique, leading to 34 to 40 point AUROC gains for each statistic in multi-domain hallucination detection scenarios.

\bibliography{example_paper}
\bibliographystyle{icml2025}

%%%%%%%%%%%%%%%%%%%%%%%%%%%%%%%%%%%%%%%%%%%%%%%%%%%%%%%%%%%%%%%%%%%%%%%%%%%%%%%
%%%%%%%%%%%%%%%%%%%%%%%%%%%%%%%%%%%%%%%%%%%%%%%%%%%%%%%%%%%%%%%%%%%%%%%%%%%%%%%
% APPENDIX
%%%%%%%%%%%%%%%%%%%%%%%%%%%%%%%%%%%%%%%%%%%%%%%%%%%%%%%%%%%%%%%%%%%%%%%%%%%%%%%
%%%%%%%%%%%%%%%%%%%%%%%%%%%%%%%%%%%%%%%%%%%%%%%%%%%%%%%%%%%%%%%%%%%%%%%%%%%%%%%
\newpage
\appendix
\onecolumn

\section{Related Work: LLM Hallucination Detection}

In this section we briefly review some of the important families of hallucination detection methods with a special emphasis on those that use the internal states of a model.

\paragraph{Retrieval-Augmented Generation:} RAG approaches address hallucinations preemptively by grounding LLM outputs in external knowledge. \citet{lewis2020retrieval} pioneered this technique by combining dense retrieval with sequence generation. Recent advances include adaptive retrieval \cite{asai2023self}, where systems dynamically determine when to retrieve information. These systems often require ground truth source data, specialized LLM training, significant infrastructure for ground truth data retrieval, and additional inference costs.

\paragraph{Consistency Methods:} Consistency-based approaches detect contradictions as hallucination indicators. Self-consistency methods such as SelfCheckGPT \cite{manakul2023selfcheckgpt} generate multiple responses and identify inconsistencies among them. More recently, \citet{chen2024inside} leverage semantic information from the internal states of LLMs to boost consistency methods.

\paragraph{Logit-based Methods:} Several hallucination detection methods leverage the logits during token generation to detect hallucinations. \citet{farquhar2024detecting} proposed to evaluate the truthfulness of LLM generations using \textit{semantic entropy}. That is, to deduplicate semantically equivalent generations and calculate the entropy of the grouped response. \citet{kadavath2022language} propose P(True), where LLMs assess the veracity of their own generations using the output logit for ``True''.

\paragraph{External Validation (Judge LLM):} External validation employs separate models or systems to assess LLM outputs. Judge LLM approaches \cite{zheng2023judging} use specialized models trained to evaluate factuality of other models' outputs. \citet{jacovi2025facts} employ an ensemble of powerful judge LLMs with sophisticated prompting techniques to check the veracity of LLM responses using ground-truth sources.

\paragraph{Detection with Internal States:} There are several approaches that exploit LLM internal representations to detect hallucinations. \citet{azaria2023internal} showed that lightweight probes can be trained with labeled data to detect hallucinations. Similar to our work, \cite{liu2024universal} showed that these probes sometimes fail to generalize between domains. \cite{liu2024universal} showed that more diverse training sets are key to probes that generalize. More recently, \citet{sriramanan2024llm} employ geometric statistics (Hidden Score and Attention Score) as detector scores for zero-resource hallucination detection. These statistics do not require ground truth and are extremely efficient to compute, but we show that they do not generalize well across domains. \citet{du2024haloscope} identify hallucinations as outliers when projecting hidden states to a lower-dimensional subspace. They use this geometric indicator to assign pseudo-labels to generated responses and subsequently train a lightweight probe on the pseudo-labeled hidden states.

\section{Dataset Details}

Our dataset has $550$ baseline correct QA pairs which are augmented with the PR template to generate hallucinations of various types. \textit{(Under-)confidence} is simulated by setting \texttt{conf\_mod} to ``probably'' (level 1), ``maybe'' (level 2), or ``not'' (level 3). \textit{Incoherence} is simulated by repeating the response $n$ times with inconsistent, incorrect answers until the last repetition in which the answer is correct. The response is repeated 2 times for level 1, 3 times for level 2, and 4 times for level 3. \textit{Incompleteness} is simulated by truncating the response early with an end-of-text token. The response is terminated after $90\%$ for level 1, $80\%$ for level 2, or $70\%$ for level 3. 

\paragraph{\texttt{math}} Our \texttt{math} domain contains $400$ integer multiplication QA pairs in the format $a \times b = \texttt{answer}$, with $a,b\in[40, 60)$. \textit{Incorrectness} is simulated by setting \texttt{answer\_offset} to a random integer between $\pm[1,9]$ (level 1), $\pm[10,99]$ (level 2), or $\pm[100,999]$ (level 3). Given a \texttt{response\_question} with $a$ and $b$, \textit{irrelevance} is simulated by sampling a new $b'\in \mathbb{Z}$ where $b' \neq b$ such that $|b' - b| < 5$ (level 1) or $5 < |b' - b| < 20$ (level 2) and inserting it into \texttt{prompt\_question}. Level 3 \textit{irrelevance} is simulated by sampling a question from another domain and using it as the \texttt{prompt\_question}.

\paragraph{\texttt{history}} Our \texttt{history} domain contains 70 QA pairs consisting of famous events and their year for three geographical regions. Each geographical region has at least two sets of ten questions each corresponding to specific timeframes. \textit{Incorrectness} is simulated by adding a random integer between $\pm[1,5]$ (level 1), $\pm[6,20]$ (level 2), or $\pm[21,50]$ (level 3). \textit{Irrelevance} is simulated by resampling a question from the same timeframe and region (level 1) or by resampling a question from a different timeframe (level 2). Level 3 \textit{irrelevance} is simulated by sampling a question from another domain and using it as the \texttt{prompt\_question}.

\paragraph{\texttt{counting}} Our \texttt{counting} domain contains 80 QA pairs of word sequences and word counts (e.g., ``\textit{word} \textit{word}, \textit{word}'' and 3). The sequences are generated by iterating through a set of eight words and repeating the word for $c \in [3, 12]$ times. \textit{Incorrectness} is simulated by adding a random integer: $\pm 1$ (level 1), $\pm 2$ (level 2), or $\pm 3$ (level 3). 
\textit{Irrelevance} is simulated by resampling a new sequence of the same word which has a count difference $|c-c'| \leq 3$ (level 1) or $|c-c'| > 3$ (level 2).
Level 3 \textit{irrelevance} is simulated by sampling a question from another domain and using it as the \texttt{prompt\_question}.

\paragraph{\texttt{all}} We include $75$ QA pairs from \texttt{math}, $70$ QA pairs from \texttt{history}, and $80$ QA pairs from \texttt{counting} into our \texttt{all} set. This ensures that the representation of the domains is relatively balanced.

\section{Additional Results}

\begin{figure*}[ht!]
    \begin{center}
        \begin{subfigure}{0.24\linewidth}
            \centering
            \includegraphics[width=\linewidth,trim=25pt 20pt 5pt 5pt, clip]{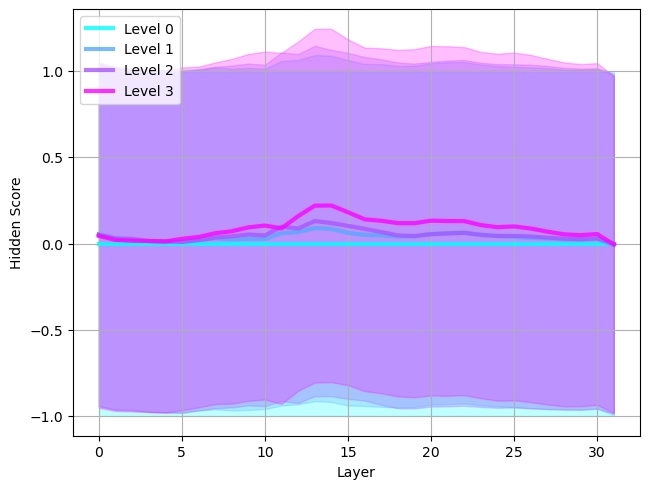}
            \caption{HS - confidence}\label{sfig:hs_conf}
        \end{subfigure}
        \hfill
        \begin{subfigure}{0.24\linewidth}
            \centering
            \includegraphics[width=\linewidth,trim=25pt 20pt 5pt 5pt, clip]{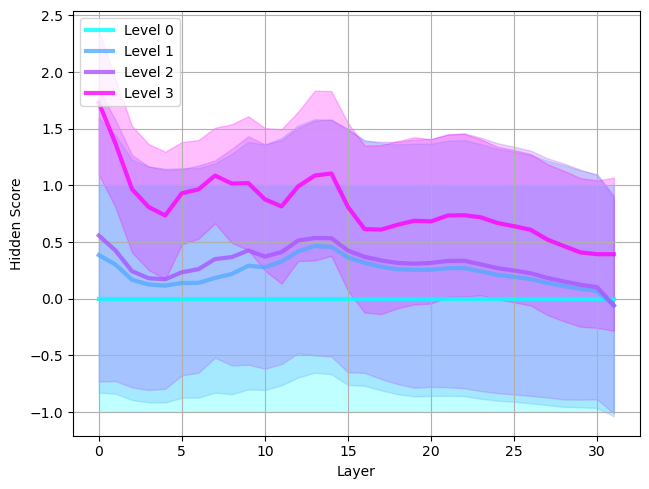}
            \caption{HS - irrelevance}\label{sfig:hs_rele}
        \end{subfigure}
        \hfill
        \begin{subfigure}{0.24\linewidth}
            \centering
            \includegraphics[width=\linewidth,trim=25pt 20pt 5pt 5pt, clip]{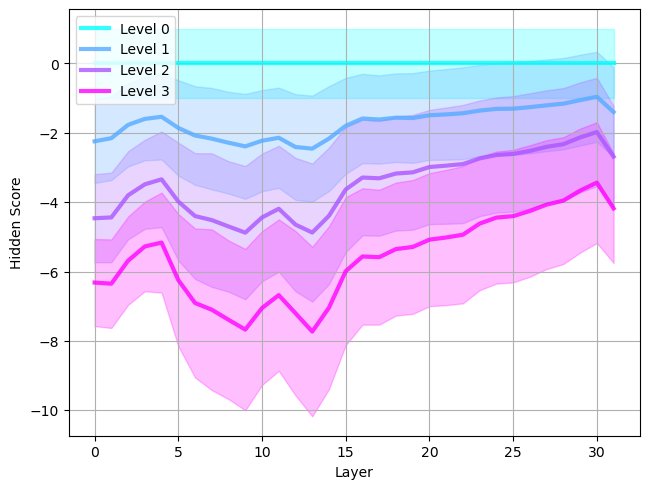}
            \caption{HS - incoherence}\label{sfig:hs_cohe}
        \end{subfigure}
        \hfill
        \begin{subfigure}{0.24\linewidth}
            \centering
            \includegraphics[width=\linewidth,trim=25pt 20pt 5pt 5pt, clip]{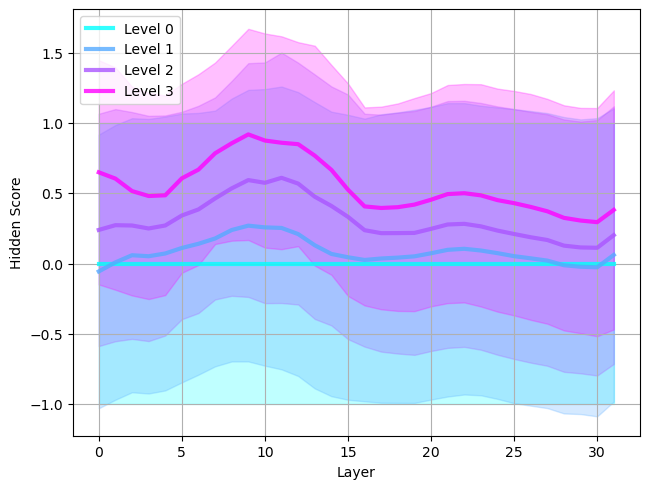}
            \caption{HS - incompleteness}\label{sfig:hs_comp}
        \end{subfigure}

        \begin{subfigure}{0.24\linewidth}
            \centering
            \includegraphics[width=\linewidth,trim=25pt 20pt 5pt 5pt, clip]{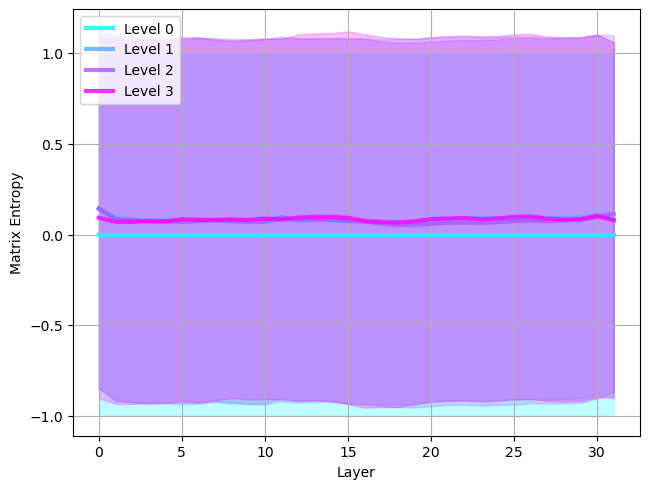}
            \caption{ME - confidence}\label{sfig:me_conf}
        \end{subfigure}
        \hfill
        \begin{subfigure}{0.24\linewidth}
            \centering
            \includegraphics[width=\linewidth,trim=25pt 20pt 5pt 5pt, clip]{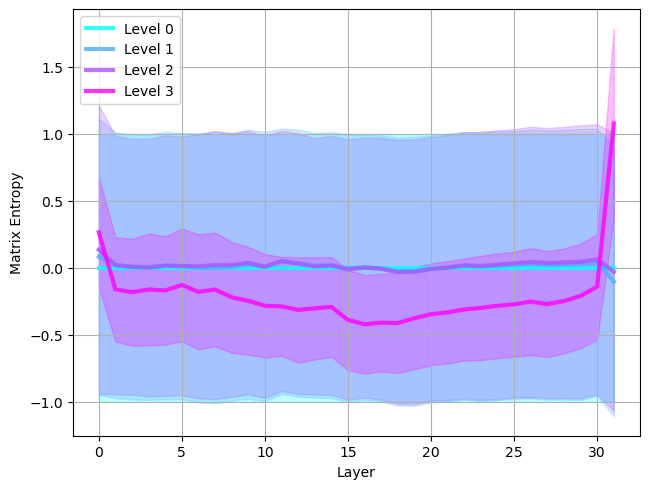}
            \caption{ME - irrelevance}\label{sfig:me_rele}
        \end{subfigure}
        \hfill
        \begin{subfigure}{0.24\linewidth}
            \centering
            \includegraphics[width=\linewidth,trim=25pt 20pt 5pt 5pt, clip]{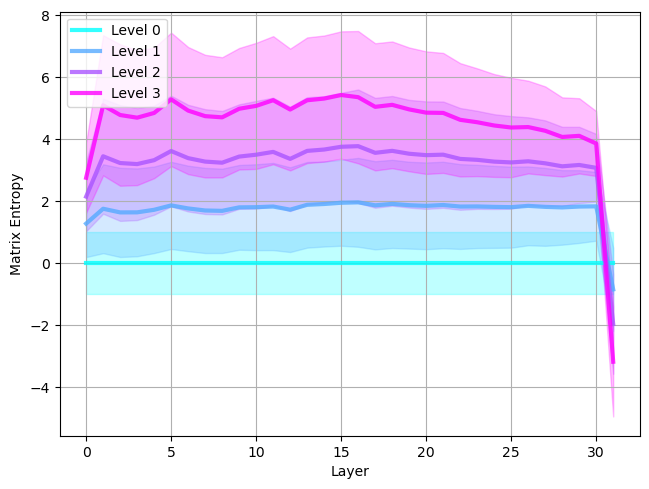}
            \caption{ME - incoherence}\label{sfig:me_cohe}
        \end{subfigure}
        \hfill
        \begin{subfigure}{0.24\linewidth}
            \centering
            \includegraphics[width=\linewidth,trim=25pt 20pt 5pt 5pt, clip]{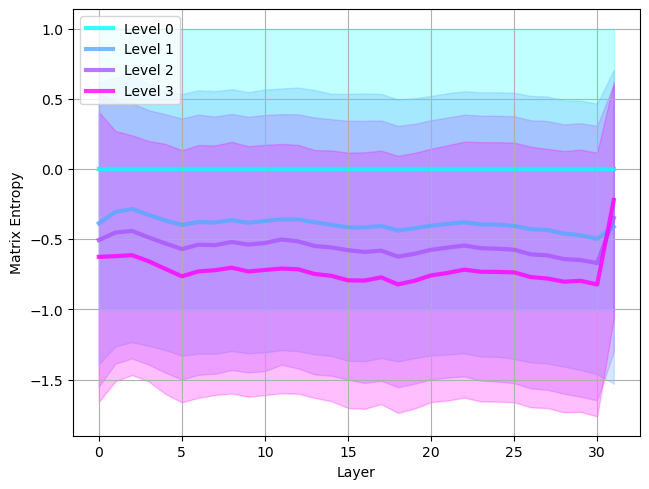}
            \caption{ME - incompleteness}\label{sfig:me_comp}
        \end{subfigure}

        \begin{subfigure}{0.24\linewidth}
            \centering
            \includegraphics[width=\linewidth,trim=25pt 5pt 5pt 5pt, clip]{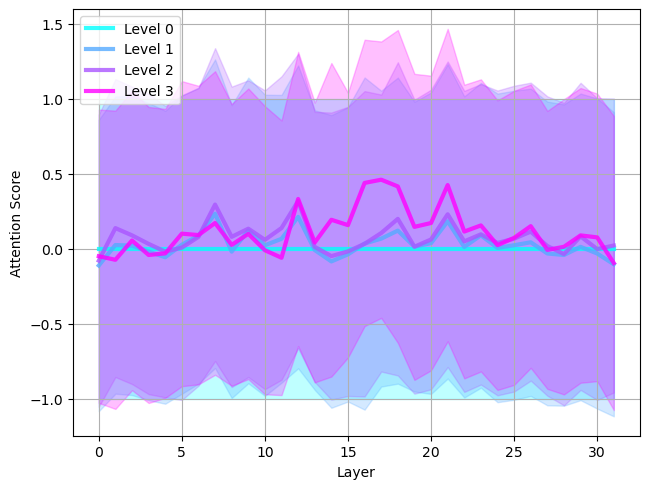}
            \caption{AS - confidence}\label{sfig:as_conf}
        \end{subfigure}
        \hfill
        \begin{subfigure}{0.24\linewidth}
            \centering
            \includegraphics[width=\linewidth,trim=25pt 5pt 5pt 5pt, clip]{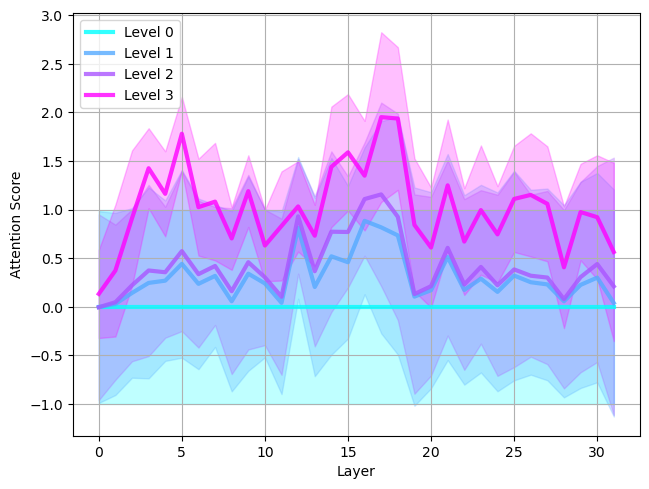}
            \caption{AS - irrelevance}\label{sfig:as_rele}
        \end{subfigure}
        \hfill
        \begin{subfigure}{0.24\linewidth}
            \centering
            \includegraphics[width=\linewidth,trim=25pt 5pt 5pt 5pt, clip]{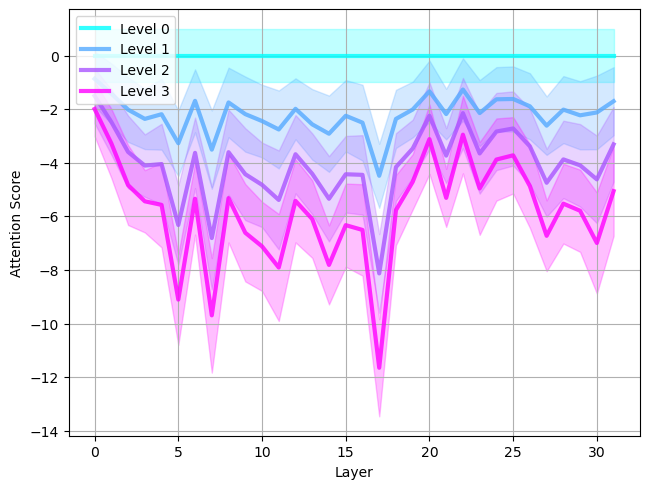}
            \caption{AS - incoherence}\label{sfig:as_cohe}
        \end{subfigure}
        \hfill
        \begin{subfigure}{0.24\linewidth}
            \centering
            \includegraphics[width=\linewidth,trim=25pt 5pt 5pt 5pt, clip]{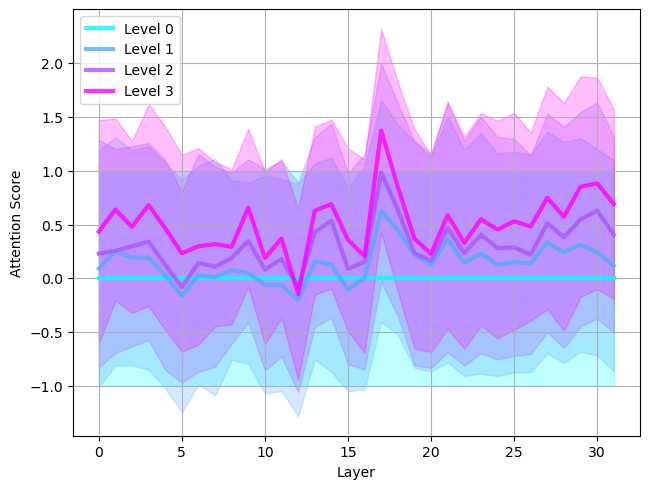}
            \caption{AS - incompleteness}\label{sfig:as_comp}
        \end{subfigure}

    \caption{The response of geometric statistics (each row) are plotted for each alternative hallucination type (each column).}
    \label{fig:axis_dataset_alt}
    \end{center}
\end{figure*}

\paragraph{Results on Alternative Hallucination Types} \cref{fig:axis_dataset_alt} depicts the distributions of responses of the geometric statistics to \textit{(under-)confidence}, \textit{irrelevance}, \textit{incoherence}, and \textit{incompleteness} for each of the $32$ layers of Llama-3.1-8B-Instruct. Each row corresponds to the responses of a single statistic (e.g., HS) to a different hallucination type on all three domains (\texttt{math}, \texttt{history}, \texttt{counting}) together. In each figure, the distribution of responses at each layer are normalized by that of the correct, base PR sequence.

\cref{sfig:hs_conf,sfig:me_conf,sfig:as_conf} show that geometric statistics are not strongly correlated with \textit{(verbal) confidence}. \cref{sfig:hs_rele,sfig:me_rele,sfig:as_rele} indicate that HS and AS respond to \textit{irrelevant} responses as hallucinations, while ME does not. However, \cref{sfig:hs_cohe,sfig:me_cohe,sfig:as_cohe} indicate that ME responds to \textit{incoherent} responses as hallucination, while HS and AS do not. We note that HS and AS are sensitive to \text{incoherence} in that this property induces significantly lower scores, but this trend is the opposite when compared to \textit{incorrectness} (hallucinations of \textit{incorrectness} are associated with higher scores). Hence, if HS or AS are used as detector scores for \textit{incorrectness} with a ``greater than or equal to'' threshold, they will not be able to detect \textit{incoherence}. \cref{sfig:hs_comp,sfig:me_comp,sfig:as_comp} show that HS and AS respond to \textit{incompleteness} as hallucination, while ME does not. Overall, these results indicate that geometric statistics are receptive to most alternative hallucination types, and that more severe hallucinations are associated with stronger responses from the statistics.

\begin{figure*}[ht]
    \begin{center}
        \begin{subfigure}{0.19\linewidth}
            \centering
            \includegraphics[width=\linewidth,trim=25pt 20pt 5pt 5pt, clip]{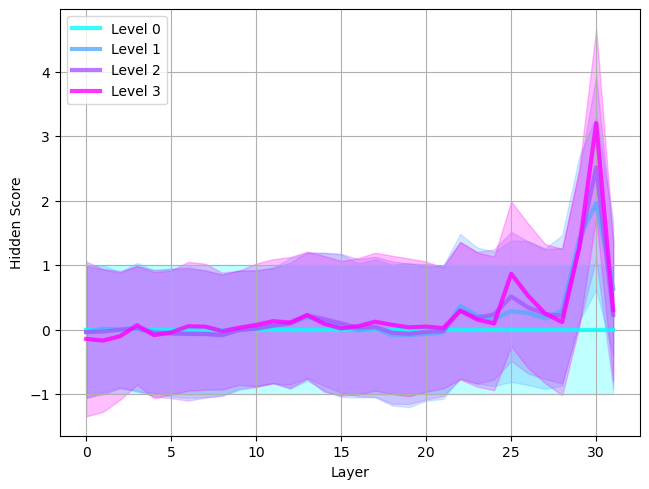}
            \caption{HS - \texttt{math}}\label{sfig:math_hs}
        \end{subfigure}
        \hfill
        \begin{subfigure}{0.19\linewidth}
            \centering
            \includegraphics[width=\linewidth,trim=25pt 20pt 5pt 5pt, clip]{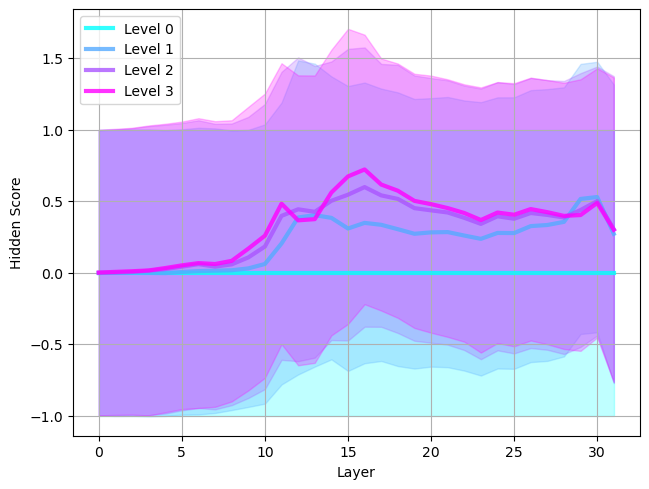}
            \caption{HS - \texttt{history}}\label{sfig:history_hs}
        \end{subfigure}
        \hfill
        \begin{subfigure}{0.19\linewidth}
            \centering
            \includegraphics[width=\linewidth,trim=25pt 20pt 5pt 5pt, clip]{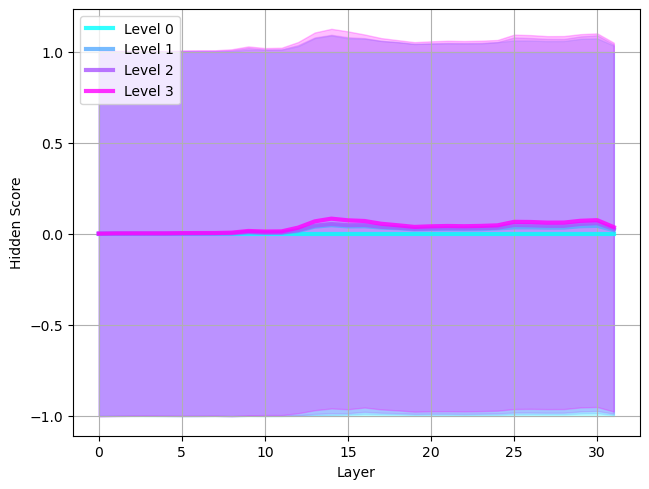}
            \caption{HS - \texttt{counting}}\label{sfig:counting_hs}
        \end{subfigure}
        \hfill
        \begin{subfigure}{0.19\linewidth}
            \centering
            \includegraphics[width=\linewidth,trim=25pt 20pt 5pt 5pt, clip]{corr_domain_hs.png}
            \caption{HS - domain comp.}\label{sfig:domain_hs}
        \end{subfigure}
        \hfill
        \begin{subfigure}{0.19\linewidth}
            \centering
            \includegraphics[width=\linewidth,trim=25pt 20pt 5pt 5pt, clip]{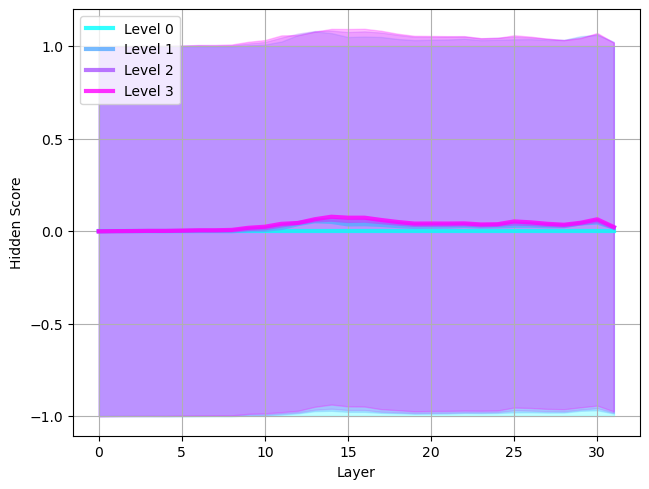}
            \caption{HS - \texttt{all}}\label{sfig:all_hs}
        \end{subfigure}

        \begin{subfigure}{0.19\linewidth}
            \centering
            \includegraphics[width=\linewidth,trim=25pt 20pt 5pt 5pt, clip]{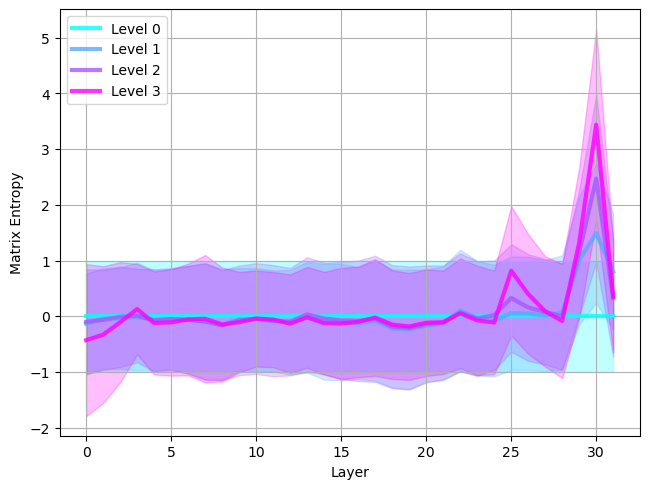}
            \caption{ME - \texttt{math}}\label{sfig:math_me}
        \end{subfigure}
        \hfill
        \begin{subfigure}{0.19\linewidth}
            \centering
            \includegraphics[width=\linewidth,trim=25pt 20pt 5pt 5pt, clip]{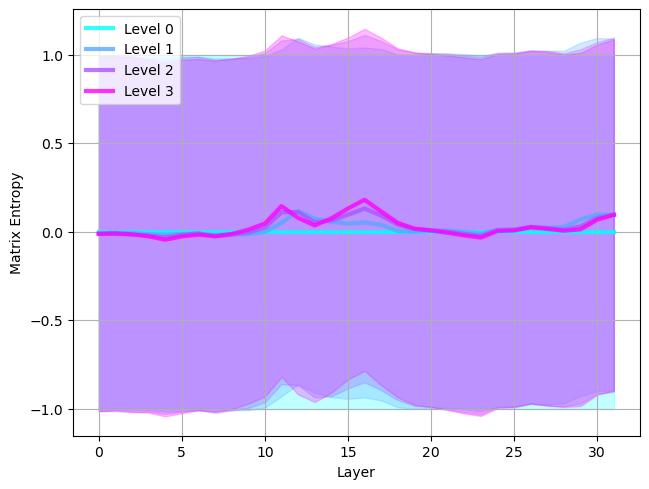}
            \caption{ME - \texttt{history}}\label{sfig:history_me}
        \end{subfigure}
        \hfill
        \begin{subfigure}{0.19\linewidth}
            \centering
            \includegraphics[width=\linewidth,trim=25pt 20pt 5pt 5pt, clip]{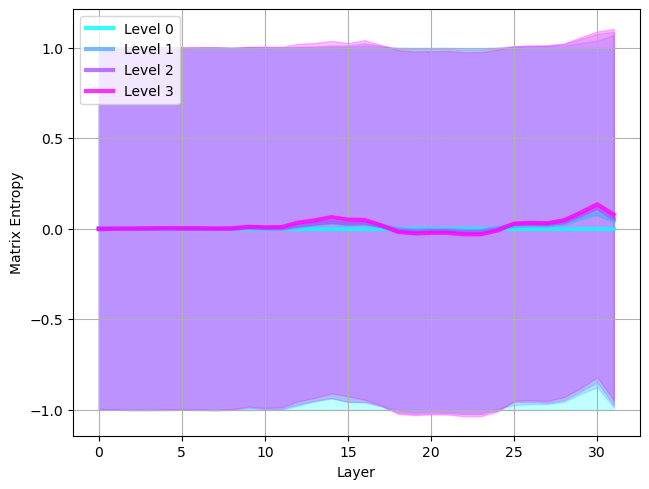}
            \caption{ME - \texttt{counting}}\label{sfig:counting_me}
        \end{subfigure}
        \hfill
        \begin{subfigure}{0.19\linewidth}
            \centering
            \includegraphics[width=\linewidth,trim=25pt 20pt 5pt 5pt, clip]{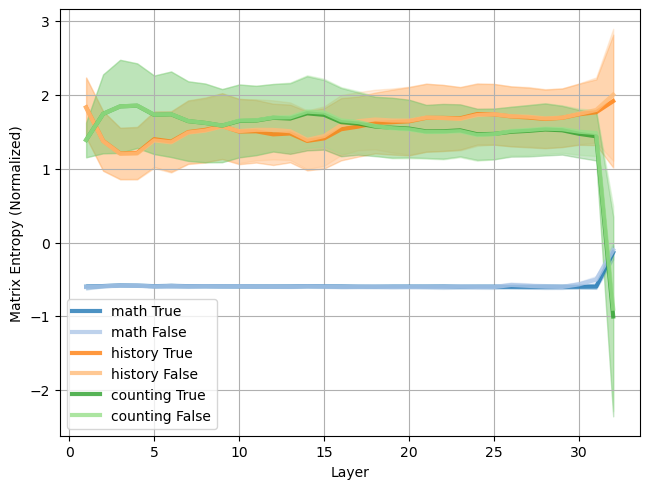}
            \caption{ME - domain comp.}\label{sfig:domain_me}
        \end{subfigure}
        \hfill
        \begin{subfigure}{0.19\linewidth}
            \centering
            \includegraphics[width=\linewidth,trim=25pt 20pt 5pt 5pt, clip]{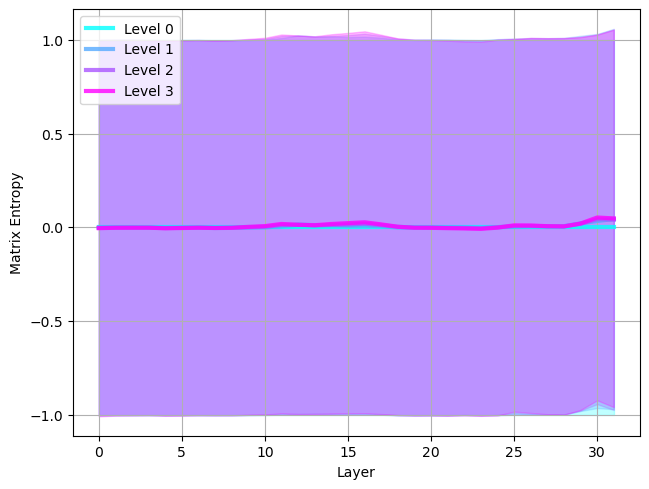}
            \caption{ME - \texttt{all}}\label{sfig:all_me}
        \end{subfigure}

        \begin{subfigure}{0.19\linewidth}
            \centering
            \includegraphics[width=\linewidth,trim=25pt 5pt 5pt 5pt, clip]{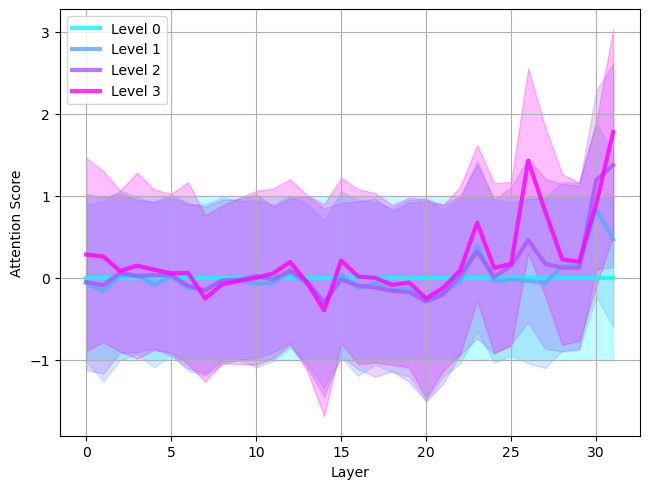}
            \caption{AS - \texttt{math}}\label{sfig:math_as}
        \end{subfigure}
        \hfill
        \begin{subfigure}{0.19\linewidth}
            \centering
            \includegraphics[width=\linewidth,trim=25pt 5pt 5pt 5pt, clip]{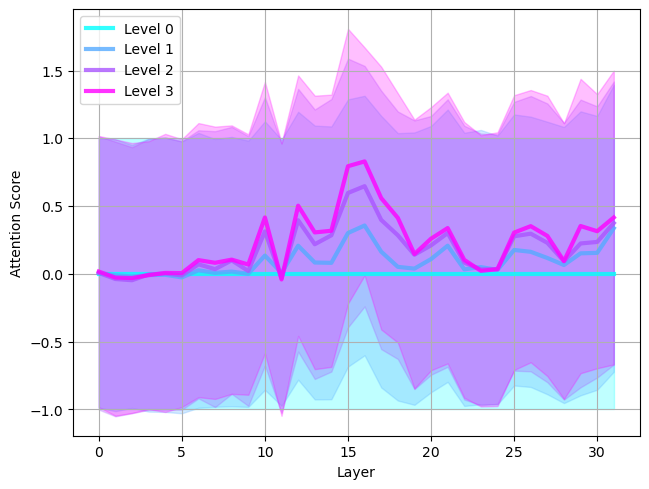}
            \caption{AS - \texttt{history}}\label{sfig:history_as}
        \end{subfigure}
        \hfill
        \begin{subfigure}{0.19\linewidth}
            \centering
            \includegraphics[width=\linewidth,trim=25pt 5pt 5pt 5pt, clip]{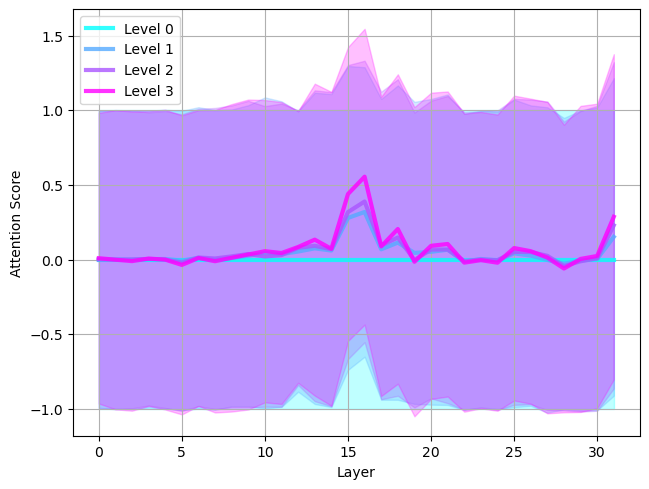}
            \caption{AS - \texttt{counting}}\label{sfig:counting_as}
        \end{subfigure}
        \hfill
        \begin{subfigure}{0.19\linewidth}
            \centering
            \includegraphics[width=\linewidth,trim=25pt 5pt 5pt 5pt, clip]{corr_domain_as.png}
            \caption{AS - domain comp.}\label{sfig:domain_as}
        \end{subfigure}
        \hfill
        \begin{subfigure}{0.19\linewidth}
            \centering
            \includegraphics[width=\linewidth,trim=25pt 5pt 5pt 5pt, clip]{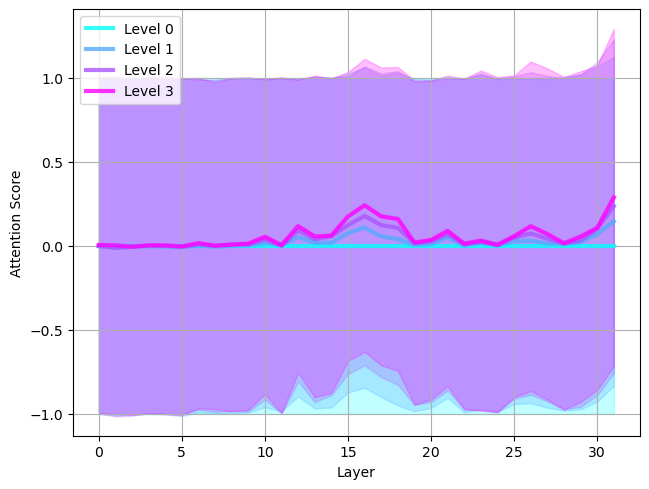}
            \caption{AS - \texttt{all}}\label{sfig:all_as}
        \end{subfigure}

    \caption{Geometric statistics (each row) are correlated with \textit{(factual)} \textit{incorrectness} on each domain (each column), motivating their use as hallucination detector scores. However, geometric statistics are impacted by domain shift, harming their cross-domain performance.}
    \label{fig:axis_data_corr_domain_shift}
    \end{center}
\end{figure*}

\paragraph{Results on Hallucinations of Facts} \cref{fig:axis_data_corr_domain_shift} depicts the distributions of geometric statistics for varying \textit{(factual) incorrectness} on the individual domains. Each row of sub-figures corresponds to a single statistic (e.g., AS).

For \texttt{math} (\cref{sfig:math_hs,sfig:math_me,sfig:math_as}), all statistics are most separable at layer $30$, near the output of the LLM. Moreover, statistics which process hidden state vectors (HS and ME) perform the best. For \texttt{history} (\cref{sfig:history_hs,sfig:history_me,sfig:history_as}), all statistics are most separable just after the midpoint of the LLM, at layer $16$. HS and AS separate \textit{factual} from \textit{non-factual} responses relatively well for \texttt{history}. For \texttt{counting}, only AS is responsive to non-factual responses. Its most separable layer for \texttt{counting} is also $16$.

\cref{sfig:domain_hs,sfig:domain_me,sfig:domain_as} depict the impact of domain shift on each of the statistics. The domain-specific distributions for each statistic are shifted and run parallel through the layers of the network. HS and ME are further impacted by the scale of the hidden states - their variance is much larger on \texttt{counting} than on \texttt{math}. These differences across domains impact detection performance on \texttt{all} subsets negatively. \cref{sfig:all_hs,sfig:all_me,sfig:all_hs} depict statistic responses across \texttt{all} domains. HS and ME show little to no separation on \texttt{all}. Of the three statistics, AS is least impacted by domain shift, leading to better performance on \texttt{all}. AS shows a small amount of separation at layers $16$ and $31$ (due to its performance on \texttt{history} and \texttt{math}, respectively).

Overall, the findings reveal that geometric statistics are impacted by domain shift, hindering their performance in more practical settings with variable domains. AS appears to be the least impacted by domain shift. We conjecture that this is due to the normalization of attention maps through the softmax function.

\begin{figure*}[ht]
    \centering
    \begin{subfigure}{0.32\linewidth}
    \centering
        \includegraphics[width=\linewidth,trim=25pt 5pt 5pt 5pt, clip]{corr_diff_domain_hs.png}
        \caption{HS-Norm - Domain Alignment}
    \end{subfigure}
    \hfill
    \begin{subfigure}{0.32\linewidth}
    \centering
        \includegraphics[width=\linewidth,trim=25pt 5pt 5pt 5pt, clip]{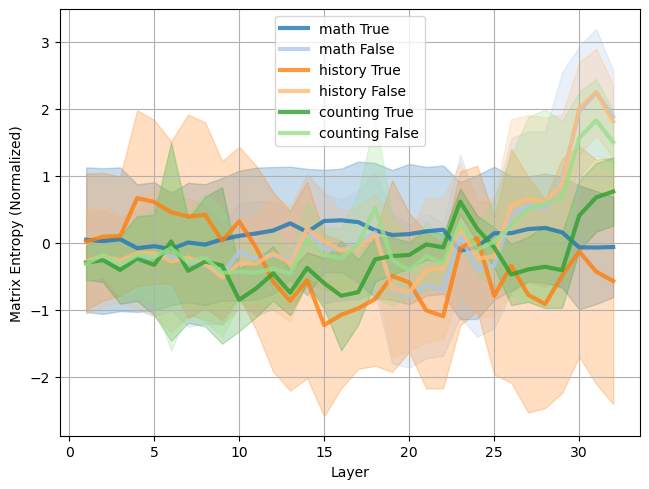}
        \caption{ME-Norm - Domain Alignment}
    \end{subfigure}
    \hfill
    \begin{subfigure}{0.32\linewidth}
    \centering
        \includegraphics[width=\linewidth,trim=25pt 5pt 5pt 5pt, clip]{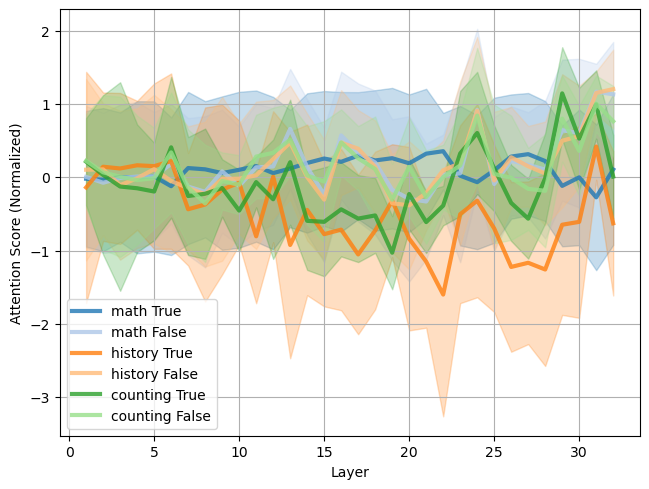}
        \caption{AS-Norm - Domain Alignment}
    \end{subfigure}
    \caption{Domain alignment of the three geometric statistics using our proposed perturbation normalization technique.}\label{fig:domain_alignment}
\end{figure*}

\begin{figure*}[ht]
    \centering
    \begin{subfigure}{0.32\linewidth}
    \centering
        \includegraphics[width=\linewidth,trim=25pt 5pt 5pt 5pt, clip]{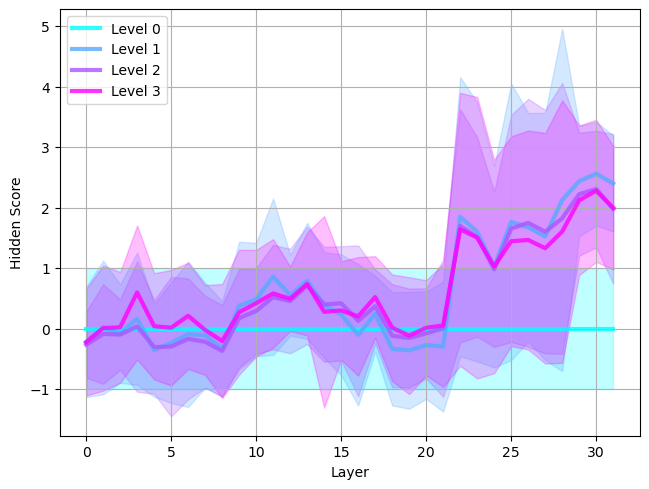}
        \caption{HS-Norm - \texttt{math}}
    \end{subfigure}
    \hfill
    \begin{subfigure}{0.32\linewidth}
    \centering
        \includegraphics[width=\linewidth,trim=25pt 5pt 5pt 5pt, clip]{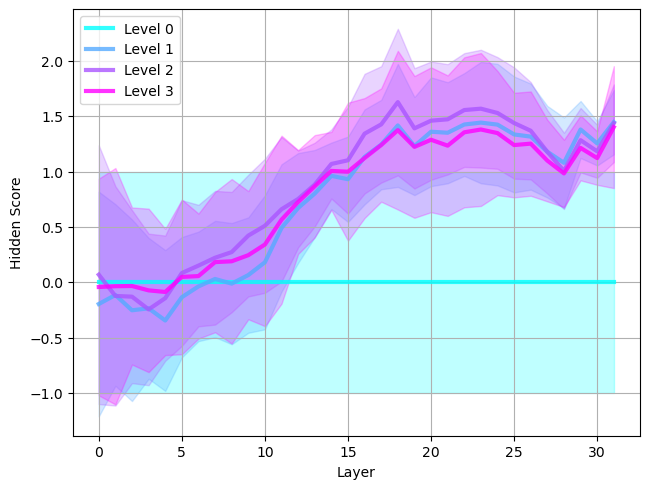}
        \caption{HS-Norm - \texttt{history}}
    \end{subfigure}
    \hfill
    \begin{subfigure}{0.32\linewidth}
    \centering
        \includegraphics[width=\linewidth,trim=25pt 5pt 5pt 5pt, clip]{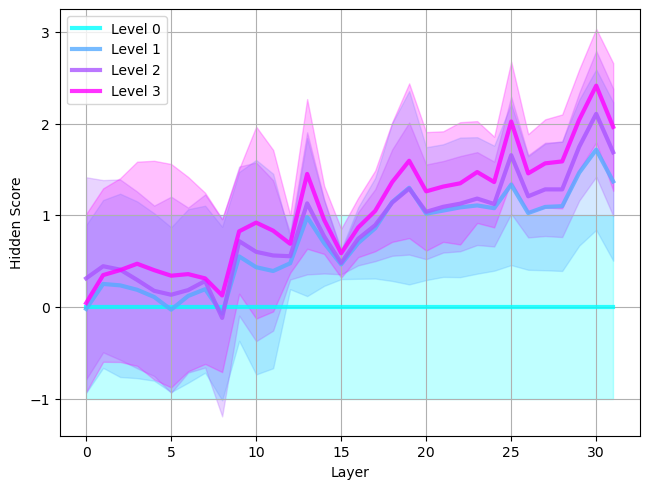}
        \caption{HS-Norm - \texttt{counting}}
    \end{subfigure}

    \begin{subfigure}{0.32\linewidth}
    \centering
        \includegraphics[width=\linewidth,trim=25pt 5pt 5pt 5pt, clip]{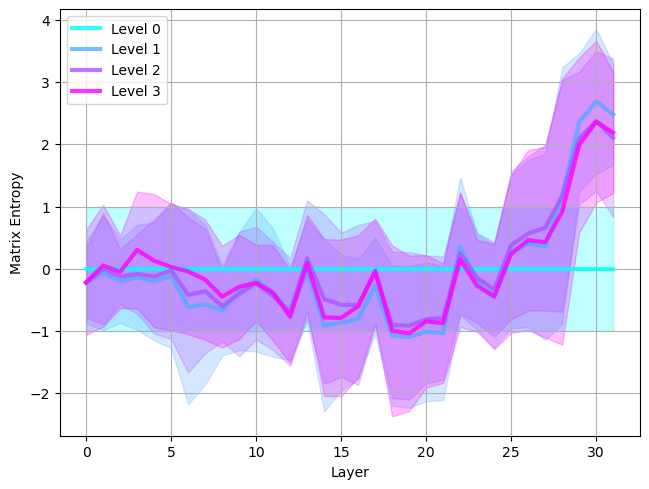}
        \caption{ME-Norm - \texttt{math}}
    \end{subfigure}
    \hfill
    \begin{subfigure}{0.32\linewidth}
    \centering
        \includegraphics[width=\linewidth,trim=25pt 5pt 5pt 5pt, clip]{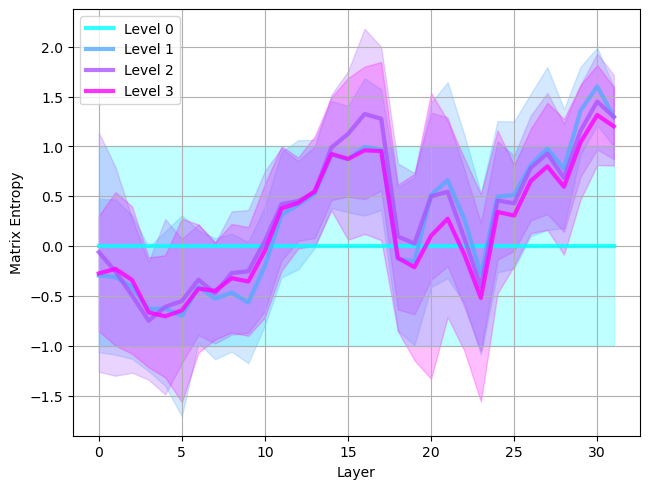}
        \caption{ME-Norm - \texttt{history}}
    \end{subfigure}
    \hfill
    \begin{subfigure}{0.32\linewidth}
    \centering
        \includegraphics[width=\linewidth,trim=25pt 5pt 5pt 5pt, clip]{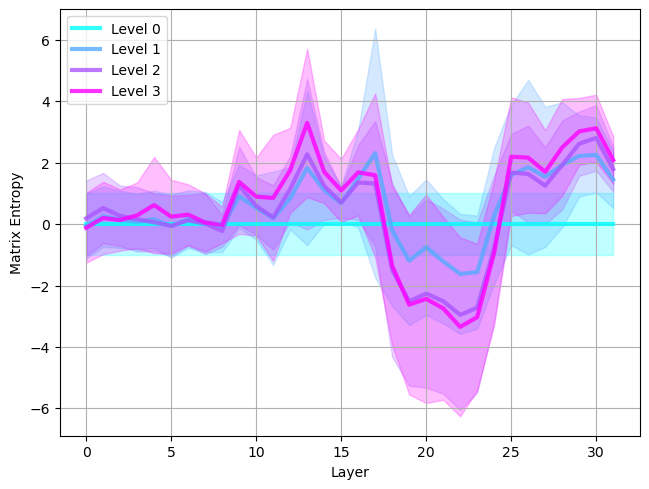}
        \caption{ME-Norm - \texttt{counting}}
    \end{subfigure}

    \begin{subfigure}{0.32\linewidth}
    \centering
        \includegraphics[width=\linewidth,trim=25pt 5pt 5pt 5pt, clip]{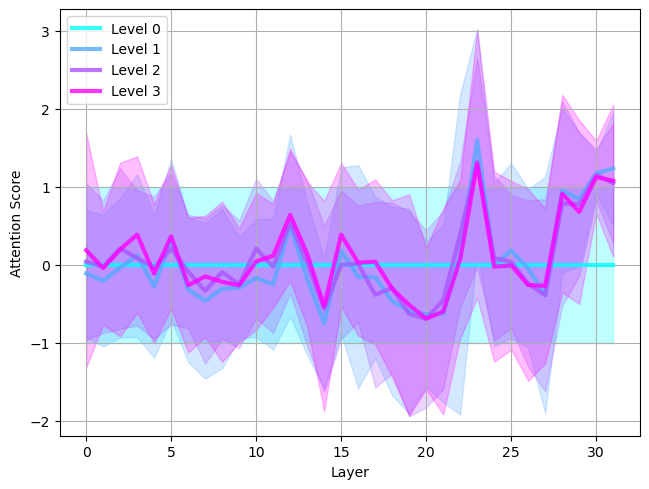}
        \caption{AS-Norm - \texttt{math}}
    \end{subfigure}
    \hfill
    \begin{subfigure}{0.32\linewidth}
    \centering
        \includegraphics[width=\linewidth,trim=25pt 5pt 5pt 5pt, clip]{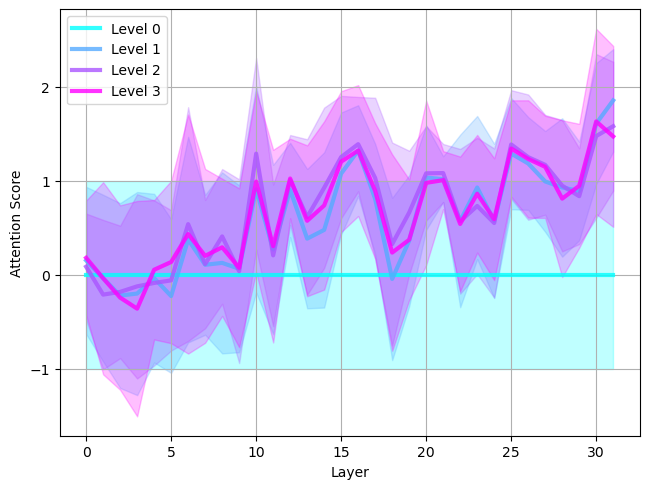}
        \caption{AS-Norm - \texttt{history}}
    \end{subfigure}
    \hfill
    \begin{subfigure}{0.32\linewidth}
    \centering
        \includegraphics[width=\linewidth,trim=25pt 5pt 5pt 5pt, clip]{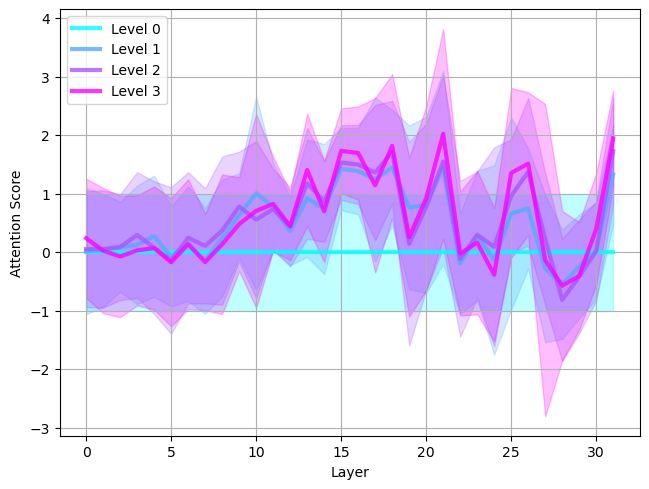}
        \caption{AS-Norm - \texttt{counting}}
    \end{subfigure}
    \caption{Geometric statistic distributions (augmented with perturbation normalization) for \textit{incorrectness} on individual domains. Perturbation normalization reduces intra-domain variance as well, enhancing single-domain detection capability.}\label{fig:domain_alignment}
\end{figure*}

%%%%%%%%%%%%%%%%%%%%%%%%%%%%%%%%%%%%%%%%%%%%%%%%%%%%%%%%%%%%%%%%%%%%%%%%%%%%%%%
%%%%%%%%%%%%%%%%%%%%%%%%%%%%%%%%%%%%%%%%%%%%%%%%%%%%%%%%%%%%%%%%%%%%%%%%%%%%%%%

\end{document}